\documentclass{article}
\usepackage{eepic,epsfig}
\usepackage{graphicx}
\usepackage{listings}
\usepackage{clrscode}
\usepackage{amsmath, amssymb, amscd, amsthm}
\usepackage{ulem}
\title{A Generic Global Constraint based on MDDs}

\author{Peter Tiedemann$^1$ \and Henrik Reif Andersen$^1$ \and Rasmus Pagh$^1$\\\\
$^ 1$ IT University of Copenhagen,\\ 
Rued Langgaards Vej 7, DK-2300 Copenhagen S, Denmark\\
$\{$petert,hra,pagh $\}$@itu.dk}
\let\oldmarginpar\marginpar

\renewcommand\marginpar[1]{\-\oldmarginpar[\raggedleft\footnotesize #1]%
{\raggedright\footnotesize #1}}

\begin{document}
\newtheorem{lemma}{Lemma}
\newtheorem{observation}[lemma]{Observation}
\newtheorem{proposition}[lemma]{Proposition}
\newtheorem{theorem}[lemma]{Theorem}
\newtheorem{definition}[lemma]{Definition}
\newtheorem{fact}[lemma]{Fact}
\newtheorem{corollary}[lemma]{Corollary}
\newtheorem{example}[lemma]{Example}
\newcommand{\comment}[1]{}
\maketitle \normalem

\begin{abstract}
  The paper suggests the use of Multi-Valued Decision Diagrams (MDDs)
  as the supporting data structure for a generic global constraint. We
  give an algorithm for maintaining generalized arc consistency (GAC)
  on this constraint that amortizes the cost of the GAC computation
  over a root-to-terminal path in the search tree. The technique used
  is an extension of the GAC algorithm for the regular language
  constraint on finite length input\cite{REGULAR-CONSTRAINT}. Our
  approach adds support for skipped variables, maintains the reduced
  property of the MDD dynamically and provides domain entailment
  detection. Finally we also show how to adapt the approach to
  constraint types that are closely related to MDDs, such as AOMDDs
  \cite{AOMDD} and Case DAGs \cite{SICTUS-PROLOG-MANUAL}.
\end{abstract}

\section{Introduction}
Constraint Programming (CP)\cite{CP-HANDBOOK} is a powerful technique
for specifying Constraint Satisfaction Problems (CSPs) based on
allowing a constraint programmer to model problems in terms of
high-level constraints. Using such \emph{global constraints} allows
easier specification of problems but also allows for faster solvers
that take advantage of the structure in the problem. The classical
approach to CSP solving is to explore the search tree of all possible
assignments to the variables in a depth-first search backtracking
manner, guided by various heuristics, until a solution is found or
proven not to exist.  One of the most basic techniques for reducing
the number of search tree nodes explored is to perform \emph{domain
propagation} at each node. In order to get as much domain propagation
as possible we wish for each constraint to remove from the variable
domains all values that cannot participate in a solution to that
constraint.  This property is known as Generalized Arc Consistency
(GAC). It is only possible to achieve GAC for some types of global
constraints in practice, as some global constraint model NP-hard
problems making GAC infeasible. The use of global constraints can
significantly reduce the total number of constraints in the model,
which again improves domain propagation if GAC or other powerful types
of consistency can be enforced. However, in typical CSPs there are
many constraints that lie outside the domain of the current Global
Constraints. Such constraints are typically represented as a
conjunction of simple logical constraints or stored in tabular form.
The former can potentially cause a massive loss in domain propagation
efficiency, while the tabular constraints typically takes up too much
space for all but the most simple constraints and for the same reason
performing domain propagation can be expensive. We aim to introduce a
new generic global constraint type for constraints on finite domains
based on the approach of \emph{compiling} an explicit, but compressed,
representation of the solution space of as many constraints as
possible. To this end we suggest the use of Multi-Valued Decision
Diagrams (MDDs). It is already known how to perform GAC in linear, or
nearly linear time in the size of the decision diagram for many types
of decision diagrams including MDD's\cite{BRYANT-BDD,AOMDD,VD-NOTE}.
However, compact as decision diagrams may be, they are still of
exponential size in the number of variables in the worst case. In
practice their size is also the main concern, even when they do not
exhibit worst case behavior. Applying the static GAC algorithms at
every step of the search is therefore likely to cause an unacceptable
overhead in many cases. To avoid this it is essential to avoid
repeating computation from scratch at each step and instead use an
algorithm that amortizes the cost of the GAC computation over a number
of domain propagation steps. In this paper we introduce such an
algorithm. In section \ref{sec:compile-vs-search} we discuss compiling
versus searching. Section \ref{sec:mdd-search} defines the type of
search we consider and the operations our constraint will support. In
section \ref{subsec:fullscan} we describe the standard GAC algorithm
for MDDs based on scanning the entire data structure and also cover
some optimizations available from related work. In section
\ref{subsec:incscan} we present a basic dynamic approach for a
simplified version of the MDD data structure based on the technique
used in \cite{REGULAR-CONSTRAINT} and contrast it to the scanning
approach covered in the preceding section. In Section
\ref{sec:long-edges} and \ref{sec:reduce} we extend the dynamic
approach to support MDDs fully as well as provide domain entailment
detection. In Section \ref{sec:construction} we discuss the issue of
constructing the MDD constraints. Finally in Section
\ref{sec:other_compilation} we show how to apply our techniques to
some other data structures that can be viewed as a compilation of the
solution space.

\subsection{Related Work}
The concept of compiling an explicit, but compact, representation of
the solution space of a set of constraints has previously been applied
to obtain backtrack-free configurators for many practical
configuration problems\cite{BDD-CONFIG}. In this case Binary Decision
Diagrams (BDDs)\cite{BRYANT-BDD} are used for representing the
solution space. However, it is well known that BDDs(and MDDs) are not
generally capable of efficiently representing constraints where the
allowable values of a variable depends on all the preceding variables
as it is then very hard to obtain good substructure sharing. This
means that prominent constraints such as the AllDifferent
constraint\cite{ALLDIFF} cannot be represented in practice unless the
number of variables is very small. Techniques have been developed for
achieving GAC in BDDs under the restriction of an external constraint,
but as we show in section \ref{sec:compile-vs-search} this technique
cannot be applied when the external constraint is the AllDifferent
constraint.  This motivates a compromise between compilation and
search, such as it is achieved by the MDD global constraint presented
in this paper.

The regular language constraint for finite sequences of variables is
introduced in \cite{REGULAR-CONSTRAINT}. It uses a DFA to represent
the valid inputs where the input is limited to be of length $n$. Since
the constraint considers a finite number of inputs, these can be
mapped to $n$ variables, and the constraint can be made Generalized
Arc Consistent according to these variables domains. To this end the
cycles in the DFA are 'unfolded' by taking advantage of the fact that
the input is of a finite length. The resulting data structure has size
$O(nd_{max}q)$ where $n$ is the number variables, $q$ the number of
states in the DFA and $d_{max}$ the size of the largest variable
domain. A GAC algorithm that amortizes the cost of the GAC computation
over root-to-leaf paths in the search tree based on this data
structure is also presented. We note that there is a strong
correspondence between the unfolded DFA and an MDD representing the
same constraint, but there are some important extra requirements on
the MDD structure which we take into account in this paper. However,
the GAC algorithm on the unfolded DFA in the regular constraint still
forms the basis of our GAC algorithm for the MDD. Below we summarize
our new contributions and highlight the differences compared to the
regular constraint.

\begin{itemize}
\item DFAs do not allow skipping inputs, even for states where the
  next input is irrelevant. Skipping input variables in this manner is
  part of the reduce steps for BDDs, and if used in MDDs requires
  alterations to the GAC algorithm. We give a modified algorithm to
  handle this. In some cases allowing the decision diagram to skip
  variables can give a significant reduction in size. A very simple
  example is a constraint specifying that the value $v$ must occur at
  least once for one of the variables $x_1, \ldots, x_n$. In an MDD
  that does not allow skipped variables(or an unfolded DFA) this
  requires $\Omega(n^2)$ nodes compared to $O(n)$ nodes if we allow
  skipped variables in the MDD.

\item BDDs are normally kept reduced during operations on the BDDs.
  This allows subsequent operations to run faster and also shows
  directly if the result is the constant true function. We present an
  approach that can dynamically reduce the MDD without resorting to
  scanning the entire live part of the data structure, and which also
  allows us to detect \emph{domain entailment}
  \cite{vanhentenryck94design}. Such entailment detection can be a
  very important property of a global constraint as it can save
  processing of the entailed constraint in all descendant search
  nodes. We note that implemented state-of-the-art CSP solvers, such
  as GeCode\cite{GECODE}, allow constraints to signal entailment in
  order to optimize the search process.

  The suggestion in \cite{REGULAR-CONSTRAINT} is to minimize the DFA
  only once at the beginning (thereby also minimizing the initial
  'unfolded' DFA) which would correspond to reducing the MDD prior to
  the search, and does not provide any form of entailment detection.
  The problem of efficient dynamic minimization is not discussed in
  \cite{REGULAR-CONSTRAINT} and would seem to require a technique
  similar to the one we present in this paper for obtaining dynamic
  reduction of the MDD constraint.

\item We adapt the GAC algorithm to operate on other decision diagram
  style data structures such as AOMDDs \cite{AOMDD} and Case DAGs
  \cite{SICTUS-PROLOG-MANUAL}.
\end{itemize}

Finally, from a practical perspective its not a good idea to first
construct a DFA and then unfold it. It is more efficient to use a BDD
package to construct an ROBDD directly, as efficient BDD packages
\cite{BUDDY,CUDD} with a focus on optimizing the construction phase
have already been developed driven by needs in formal verification
\cite{VLSI}. Specifically the use of BDDs for the construction gives
access to the extensive work done on variable ordering (see for
example \cite{FORCE,SCATTER, SIFTING}) for BDDs.  Once an ROBDD is
constructed it can then easily be converted into the desired MDD.

Another related result is \cite{ADHOC-GAC} in which it is discussed
how to maintain Generalized Arc Consistency in a binary decision
diagram(BDD) when using the BDD as a global constraint in a CSP. The
solution approach suggested in \cite{ADHOC-GAC} is intended for
smaller constraints with a small scope, and is only presented for the
case of binary variables. Their technique differs from the
straightforward scanning technique by using shared good/no-good
recording and a simple cut-off technique to (in some cases) reduce the
amount of nodes visited in a scan. Their technique can be adapted for
non-binary variables, but good/no-good recording becomes useless as the
scope of the MDD constraint increase, and the cut-off technique lose
merit if we wish to reduce the MDD dynamically or have domains that
are even slightly larger than 2. Hence their techniques do not apply
when the intention is to collect as many small constraints as possible
into one global MDD. We discuss the direct adaption of their technique
to non-binary variables below and compare it to our approach in
section \ref{subsec:fullscan}.

An entirely different approach is suggested in \cite{BOX-SWEEP,BCC}
which considers representing constraints as a disjunction of
geometrical constraints(boxes and triangles in \cite{BCC} and just
boxes in \cite{BOX-SWEEP}. The experiments in \cite{BCC} gives a
comparison with the case constraint\cite{SICTUS-PROLOG-MANUAL} which
is implemented using what is essentially an MDD, called a case DAG,
where edges represents an interval of values. They also provide
comparisons with using a case DAG directly along with a simple GAC
algorithm that treats the DAG as a tree. The experiments show that the
naive DAG approach is slower than the box and triangle approach, but
the case implementation (which most likely uses a DFS scanning
approach) is still faster. In section \ref{subsec:interval-edges} we
extend our dynamic approach to support the case DAG, most likely
increasing the advantage over the box and triangle approach.

\subsection{Notation}
In this paper we consider a CSP problem $\textit{CP}(X,D,F)$, where $X
= \{ x_1, \ldots , x_n \}$ is the set of variables, $F$ the set of
constraints and $D = \{ D_1, \ldots, D_n\}$ is the multi-set of
variable domains, such that the domain of a variable $x_i$ is $D_i$.
When discussing a backtracking search we will use $D_i$ to refer to
the \emph{currently} allowed domain values for $x_i$, and $D_i^0$ to
denote the original domains. We use $d_i = |D_i^0|$ to denote the size
of domains and $d_{max} = \max \{ d_i | x_i \in X \}$ to denote the
largest domain (for simplifying complexity analysis). We use
$scope(\{F_1, \ldots , F_j \})$ and $scope(F_j)\subseteq X$ to denote
the variable scope of a set of constraints and a single constraint
respectively.

A \emph{single assignment} $a$ is a pair $(x_i,v)$ where $x_i \in X$
and $v \in D^0_i$. The assignment $a$ is said to have support in a
constraint $F_k$, iff there exists a solution to $F_k$ where $x_i$ is
assigned $v$. If a single assignment ($x_i,v$) has support in a
constraint $F_j$, $v$ is said to be in the valid domain for $x_i$,
denoted $VD_i(F_j)$. If for all variables $x_i$ and a constraint $F_j$
it is the case that $D_i = VD_i(F_j)$ then $F_j$ is said to fulfill
the property of \emph{Generalized Arc Consistency} (GAC). A
\emph{partial assignment} $\rho$ is a set of single assignments to
distinct variables, and a \emph{full assignment} is a partial
assignment that assigns all variables.

\subsection{The MDD data structure}
Below we give the definition of the MDD data structure. We will then
present the valid domains algorithm and show how to store the MDD to
support the suggested algorithms.

\begin{definition}[Ordered Multi-Valued Decision Diagram (OMDD)]
An Ordered Multi-Valued Decision Diagram (OMDD or just MDD) for a CSP
\textit{CP} is a layered Directed Acyclic MultiGraph $G(V,E)$ with up
to $n+1$ layers. Each node $u$ has a label $l(u) \in \{1, \ldots,
n+1\}$ corresponding to the layer in which the node is placed, and
each edge $e$ outgoing from layer $i$ has a label $v(e) \in D_i$.
Furthermore we use $s(e)$ and $d(e)$ to denote the source and
destination layer of each edge $e$ respectively.

The following restrictions apply:
\begin{itemize}
\item There is exactly one node $u$ such that $l(u) = \min\{ l(q) \mid
  q \in V\}$ denoted \textit{root}.
\item There is exactly one node $u$ such that $l(u) = n+1$ denoted
\textit{terminal}.
\item For any node $u$, all outgoing edges from $u$ have distinct
labels.
\item All nodes except $terminal$ has at least one outgoing edge.
\item For all $e \in E$ it is the case that $s(e) < l(e)$.
\end{itemize}

A full assignment $\rho$ is a solution to a given MDD iff there exists
a path $Q = (e_1,\ldots, e_j)$ from \textit{root} to \textit{terminal}
such that for each $(x_i,v) \in \rho$ there exists an edge $e \in Q$
such that $l(e) = i$ and $v(e) = v$ or $s(e) < i < d(e)$.
\end{definition}

\begin{figure}
\includegraphics[height=10cm]{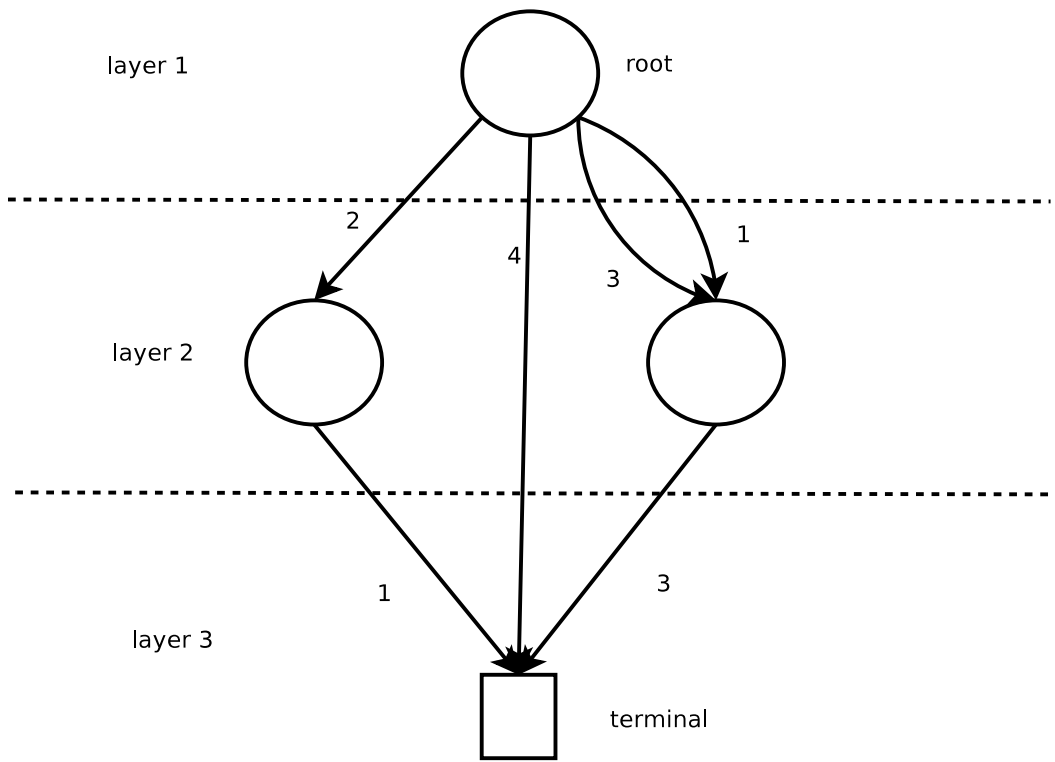}
\caption{The above figure shows an example MDD that is fully
  reduced. Assuming $D_1= \{ 1,2,3,4\}$ and $D_2 = \{ 1,2,3\}$, it
  represents the binary constraint with solutions \{ (1,3),(2,1),
  (3,3),(4,1),(4,2),(4,3) \}.}
\label{fig:MDD}
\end{figure}

We will use $V_i = \{ u \in V \mid l(u) = i\}$ to denote the nodes of
layer $i$ and $E_i = \{ e \in E \mid s(e) = i\}$ to denote the set of
edges originating from layer $i$. Furthermore we define $P_u = \{
(p,v) \mid \exists e \in E : v(e) = v \land s(e) = p \land d(e) = u\}$
and $C_u = \{ (c,v) \mid \exists e \in E : v(e) = v \land s(e) = u
\land d(e) = c \}$. That is, $P_u$ corresponds to the incoming edges
to $u$, and $C_u$ corresponds to the outgoing edges of $u$.

\begin{definition}[Reduced OMDD]
  An MDD is called \emph{uniqueness reduced} iff for any two distinct
  nodes $u_1,u_2$ at any layer $i$ it is the case that $C_{u_1} \not =
  C_{u_2}$.

  If it is furthermore the case for all layers $i \in \{1, \ldots,n\}$
  that no node $u_1$ in layer $i$ exists with $d_{i}$ outgoing edges
  to the same node $u_2$, the MDD is said to be \emph{fully reduced}.
\end{definition}

The above definitions are just the straightforward extension of the
similar properties of BDDs\cite{BRYANT-BDD}. Fully reduced MDDs retain
the canonicity property of reduced BDDs, that is there is exactly one
fully reduced MDD for each Boolean constraint on $n$ discrete domain
variables. An example MDD is shown in Figure \ref{fig:MDD}

\section{Compiling vs. searching}\label{sec:compile-vs-search}
Consider the problem of \emph{compiling} the set of all possible
solutions to a CSP. By compiling we mean computing an explicit, but
compressed, representation of the set of solutions to the CSP, such
that evaluation of assignments in time polynomial in the size of the
representation is supported. While compiling the solution space is
obviously harder than finding a single solution to the constraint set,
this approach has been used successfully with BDDs for verification of
circuits (as described above) and for interactive
configuration\cite{BDD-CONFIG}. That is, in certain scenarios it is
possible to compile the \emph{entire} solution space of a CSP problem
which can be viewed as obtaining one huge global constraint upon which
GAC can be enforced. There are of course many global constraint types
which by themselves result in a decision diagram that is too large to
handle. One of the most prominent is the AllDifferent constraint. As
there are many practical applications where an AllDifferent constraint
plays a crucial role (such as layout/placement problems, which also
frequently occur in configuration problems) it seems obvious that a
search approach is appropriate.

However, in a recent result \cite{COST-CONFIG} it was shown how to
perform valid domains computation on a BDD under the further
restriction of a separate linear constraint(which if included in the
BDD might have produced an exponential blow-up in size) in polynomial
time in the size of the BDD. This allows efficient cost configuration
for a restricted class of cost functions. That is, in some cases it is
possible to \emph{filter} the valid domains computation enforcing
additional constraints without encoding the additional constraints
directly in the decision diagram. It is obvious to consider whether or
not a similar approach can be taken for the problematic AllDifferent
constraint. However, as we show below, performing valid domains
computation under the restriction of an AllDifferent constraint in
polynomial time in the size of the BDD implies that P=NP.

\begin{theorem}
  If an algorithm exists for checking satisfiability of the
  conjunction of an MDD and an AllDifferent constraint in time
  polynomial in the size of the MDD then P = NP.
\end{theorem}
\begin{proof}
  We consider the Hamiltonian Path problem\cite{HAMPATH} on an
  undirected graph $G(V_H,E_H)$. We will show that the Hamiltonian
  Path problem can be expressed as an AllDifferent constraint
  conjoined with an MDD of size polynomial in the number of nodes in
  the input graph. Therefore the existence of a polynomial time
  algorithm deciding whether or not an MDD contains a solution that
  satisfies an AllDifferent constraint implies P = NP.

  We model the Hamiltonian path  problem with $n = |V_H|$ variables of
  domain size $n$. The value  of the $i$th variable corresponds to the
  $i$th  node  visited in  the  Hamiltonian path.  We  use  an MDD  to
  represent the  \emph{N-Walk constraint}, which  restricts the values
  of the $n$ variables to represent  a valid $n$ node long walk of the
  graph.  That is, the  edge labels  of a  path from  \textit{root} to
  \textit{terminal} in the MDD gives a  valid walk of $n$ steps in the
  graph  $G_H$. An  example is  provided in  Figure \ref{fig:hampath}.
  When  combined   with  an  AllDifferent  constraint   over  the  $n$
  variables, we have a representation of the Hamiltonian Path problem.

\begin{figure}\label{fig:hampath}
\includegraphics[height=10cm]{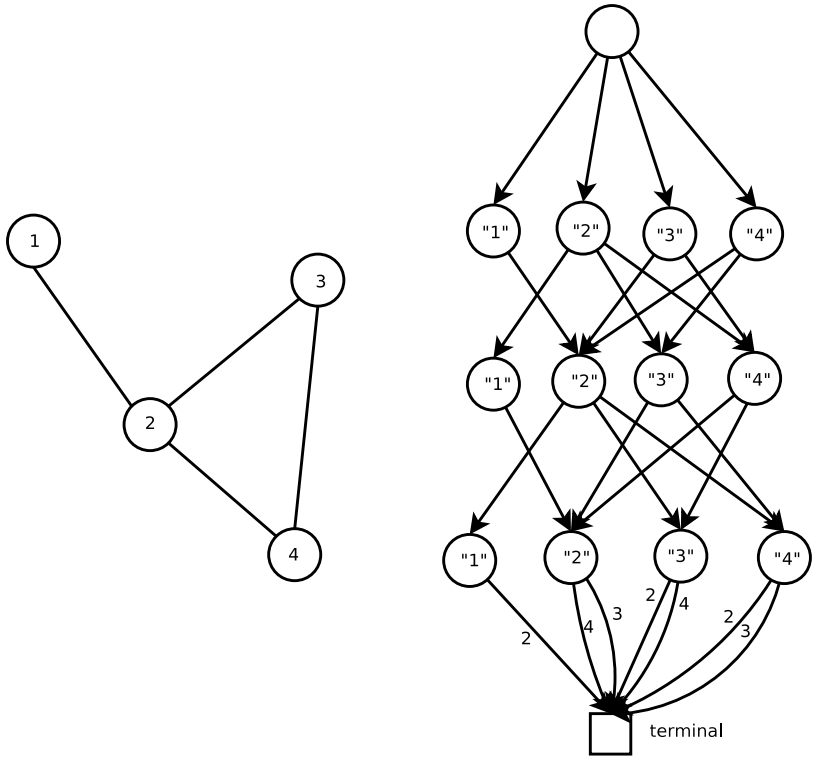}
\caption{An example of an MDD encoding an N-Walk constraint. On the
  left is shown an example graph, and on the right the corresponding
  MDD. In order to avoid clutter some edge labels have been left out
  in the MDD and instead edges leading to a node labelled for example
  ``1'',''2'',''3'' or ``4'' correspond to edge labels 1,2,3 and 4
  respectively (this type of labelling is only possible because all
  edges leading to the same MDD node always have the same label in the
  N-Walk constraint). The root represents the choice point for the
  start of the walk, and nodes in layer 1 represents the choice point
  for the second node in the walk. }
\end{figure}

  There are only $O(n^2)$ different states in the N-Walk constraint,
  as the valid choices for the current variable depend only on the
  value of the preceding variable (the node in $G_H$ we are at now)
  and the number of variables already assigned (how many nodes have
  been visited so far). Hence the MDD representation is obviously
  polynomial, having at most $O(n^2)$ nodes and at most $O(n^3)$
  edges, assuming it is uniqueness reduced.
\end{proof}

From this result we get a strong motivation for settling on a search
strategy. One could argue that alternative compiled data structures
providing valid domains computation under the restriction of an
AllDifferent constraint in polynomial time of its size could exist.
However, it is obvious from the above result that any such data
structure must require super polynomial construction time in the
worst-case when encoding the simple N-Walk constraint (unless $P=NP$).

\section{Searching with an MDD}\label{sec:mdd-search}
In this paper we consider a backtracking search for a solution to a
conjunction of constraints at least one of which is an MDD. To
simplify the complexity analysis, we assume that the search branches
on the domain values of each variable in some specified order and that
full domain propagation takes place after each branching. The process
of branching and performing full domain propagation we will refer to
as a \emph{phase}. For the proposed constraint we refer to performing
domain propagation on the MDD, as a single \emph{step}. As such, one
phase may contain many steps depending on how many iterations it takes
until none of the constraints are able to remove any further domain
values.

In order to be useful in this type of CSP search, an implementation of
the MDD constraint needs to supply the following functionality:

\begin{itemize}
\item \proc{Assign}($x,v$)
\item \proc{Remove}($(x_1,v_1),\ldots, (x_k,v_k)$)
\item \proc{Backtrack}()
\end{itemize}

The \proc{Assign} operation restricts the valid domain of $x$ to $v$
and is used to perform branchings. The \proc{Remove} operation
corresponds to domain restrictions occurring due to domain propagation
in the other constraints. The \proc{Backtrack} operations undoes the
last \proc{Assign} operation and all \proc{Remove} operations that has
occurred since, effectively backtracking one phase in the search tree.

For the implementation of \proc{Backtrack} we will simply push data
structure changes on a stack, so that they can be reversed easily when
a backtrack is requested. This very simple method ensures that a
backtrack can be performed in time linear in the number of data
structure changes made in the last step. For all the dynamic data
structures considered in this paper the space used for this undo stack
will be asymptotically bounded by the time used over a
root-to-terminal path in the search tree. Furthermore, in all the
cases studied in this paper, \proc{Assign}($x,v$) is just as
efficiently implemented as a single call to \proc{Remove}($ \{
(x_i,v_j) \mid v_j \in D_i \setminus \{ v \} \}$). Therefore we will
only discuss implementation of \proc{Remove}.

\section{Calculating the change in valid domains}
In this section we consider two different approaches for determining
which values are lost from the valid domains when applying
restrictions of the form $x_i \not = v$. A crucial element in
computing the valid domains is that of a \emph{supporting edge}. An
edge $e$ \emph{supports} a single assignment $(x_i,v)$, if $s(e) = i$
and $v(e) = v$ or $s(e) < i < d(e)$.  Note that the existence of an
edge supporting a give assignment implies that the assignment is part
of the valid domain for the corresponding variable.

The first approach to maintaining valid domains we cover is the
straight forward scanning approach that builds the valid domains from
scratch by scanning and finding all supporting edges. In addition to
this we discuss the direct adaptation of some optimization techniques
from \cite{ADHOC-GAC}. The second approach is the dynamic technique we
suggest based on \cite{REGULAR-CONSTRAINT}, which instead relies on
tracking the loss of supporting edges.

Finally, to ease presentation we will at first assume that the OMDD we
operate on is initially Uniqueness Reduced, but not fully reduced, in
fact we will assume that all outgoing nodes from a node in layer $i$
lead to nodes in layer $i+1$ (and hence also $l(root) = 1$). In
section \ref{sec:long-edges} we show how to handle a fully reduced
OMDD.

\subsection{The full scan algorithm}\label{subsec:fullscan}
We will now present the standard scanning algorithm for computing the
valid domains from scratch. It scans the MDD for supporting edges and
stores the assignments that they support. This is done in a DFS manner
deleting encountered edges that correspond to disallowed assignments.
A node $u$ is said to \emph{die} when it can no longer participate in
any solution, and is otherwise said to be \emph{alive}. During the DFS
search nodes that are not already known to be dead or alive are
searched recursively, and as soon as a valid path to \textit{terminal}
is found they are marked as being alive. A domain value is added to
the valid domains if an edge with a live end point corresponding to
this value is visited.  The pseudo-code for this approach is shown in
Figure \ref{code:fullscan}.

\begin{figure}
\label{code:fullscan}
\begin{codebox}
\Procname{\proc{RemoveScan}$(R)$}
\li \proc{global} $live \gets \emptyset$ \RComment{Set of nodes found to be alive}
\li \proc{global} $dead \gets \emptyset$ \RComment{Set of nodes found to be (newly) dead}
\li \For $i$ st. $x_i \in X$
\li \Do $D_i \gets \emptyset$ 
\End
\li $rootAlive \gets \proc{ScanRecursive}(root,R)$
\li \If $rootAlive = false$
\li \Then \Return Constraint failed
\End
\end{codebox}
\begin{codebox}
\Procname{\proc{ScanRecursive}($u,R$)}
\li \If $u \in live$ 
\li \Then \Return $true$
\li \ElseIf $u \in dead$
\li \Then \Return $false$
\End
\li $alive \gets false$ \RComment{$u$ is dead unless support is found}
\li \For $(c,v) \in C_u$ \RComment{Outgoing edge from $u$ to $c$} 
\li  \Do \If $c \in dead \lor (x_{l(u)},v) \in R$ 
\li  \Then $C_u \gets C_u \setminus \{(c,v)\}$ \RComment{Edge is dead}
\li  \ElseIf $\proc{ScanRecursive}(c,R)$
\li  \Then $alive \gets true$
\li  $D_{l(u)} \gets D_{l(u)} \cup \{ v \}$ \RComment{Add to valid domain}
\li  \Else $C_u \gets C_u \setminus \{(c,v)\}$ \RComment{Edge is dead}

\End 
\End 

\li \If alive
\li \Then $live \gets live \cup \{u \}$
\li \Else $dead \gets dead \cup \{u \}$
\End
\li \Return $alive$
\end{codebox}
\caption{The above pseudo-code shows the scanning approach for the
valid domains computation. The set $R$, i.e. the collection of
arguments to \proc{Remove}, consists of the pairs $(x_i,v)$ that have
been disallowed by other constraints since the last step. The set
$dead$ consists of the nodes that have been found to be newly dead,
while $live$ is the set of nodes that have been found to be alive in
this step. The decremental updates to the outgoing edges $C_u$ can be
implemented as marking in practice.}

\end{figure}
\subsubsection{Two ways for nodes to perish}
In order to analyze the complexity of the scanning algorithm it is
important to distinguish between two different causes of a node $u$
dying. Firstly, all the parents of $u$ can lose their edge leading to
$u$. If so, there is no longer any path from $root$ to $terminal$
through $u$, so $u$ can no longer be part of a solution. We will refer
to this as a \emph{NoReference} node death. Secondly, a node can lose
all its outgoing edges, so that $root$ can never be reached from it,
which we will call a \emph{NoValue} node death. Similarly an edge dies
if one of its end points die.

We will denote the set of live edges after the $i$'th step as
$E^i_{liv}$. The initial edge set preceding the first step is
$E^0$. Furthermore we use $E^i_{ref}$ to refer to the edges lost in
step $i$ due to a \emph{NoReference} node death. Similarly we will use
$E^i_{val}$ to denote the set of edges that perished in step $i$ due
to a $NoValue$ death. Note that an edge can appear in both $E^i_{ref}$
and $E^i_{val}$.

\begin{lemma} \label{fullscan_complexity}
In step $t$ \proc{RemoveScan} traverses $|E^t_{liv}| + |E^t_{val}
\setminus E^t_{ref}|$ edges.
\end{lemma}
\begin{proof}
We note that \proc{ScanRecursive} when invoked on the root will visit
all edges once except those that lie in a part of the MDD that cannot
be accessed any more due to the new restriction in $R$. If an edge $e$
is inaccessible it must by definition at least die from
\emph{NoReference}, i.e. $e \in E^t_{ref}$. It follows that edges that
die only from \emph{NoValue} must be accessible and therefore will be
traversed.
\end{proof}

\subsubsection{Good/No-good recording}
The use of good/no-good recording to assist the scan algorithm during
search is introduced in \cite{ADHOC-GAC}. The technique is extremely
simple, relying on recording the current partial assignment projected
on the scope of the constraint in question as a no-good when a
backtrack is needed. If a partial assignment occurs that matches a
previous failed partial assignment on the scope of the constraint, the
constraint will know it has failed. This can also be done for the
cases where no domains change, resulting in a 'good'
recording. Furthermore, the stored no-goods can be used by identical
constraints defined on different scopes.

Obviously, the no-good recording is less useful if the constraints
have a large scope, as the number of times the stored no-goods can
potentially be used decreases exponentially as the scope increases.

\subsubsection{$\Delta$-cutoff}
This technique was suggested in \cite{ADHOC-GAC} for use in BDDs, and
consists of the following: While scanning we maintain the largest
index $\Delta$ such that the currently discovered valid values in the
domains of $x_{ \Delta }, \ldots x_n$ are the same as the valid values
in the previous step. Should we at any time be scanning a node $v$
which has already been found to be alive( has at least one outgoing
live edge to a live node), we can neglect to recursively scan its
other children if $l(u) \ge \Delta$.

We expect $\Delta$-cutoff to be able to perform large cut-offs in a
constraint with binary variable domains, but as the domain size
increases, it will take much more scanning before a large continuous
interval of variables find all the values they were previously
allowed.
 
Furthermore, consider a node $s$ in the search tree. It might be the
case that a cut-off is made that would otherwise have pruned a number
of edges. As further restrictions are applied it is highly likely that
we at some descendant search nodes of $s$ will be required to scan the
part of the structure that was previously neglected. As there can be a
large number of descendant search nodes the $\Delta$-cutoff might
actually result in an non-constant factor performance decrease.

\section{A New dynamic approach}\label{subsec:incscan}
In this section we present a dynamic algorithm for maintaining valid
domains, based on tracking the loss of supporting edges to avoid
recomputing the valid domains from scratch in every step. It follows
the technique presented in \cite{REGULAR-CONSTRAINT}, just applied to
the MDD instead of the unfolded DFA. In the following sections we then
extend the algorithm to handle fully reduced MDDs and add support for
dynamically reducing the MDD, enabling us to deliver domain entailment
detection.

\subsection{Support lists}
In order to avoid re-doing unnecessary work we track the set of
supporting edges by storing a set of sets $S$, such that for every
possible single assignment $(x_i,v)$ where $x_i \in X$ and $v \in D_i$
there exists a set $s_{i,v} \in S$ containing all the nodes and the
corresponding edges that gives support to the single assignment
$(x_i,v)$. As we will see below $S$ can easily be maintained. In
maintaining $S$ we learn immediately when a single assignment no
longer has support, as the corresponding $s_{i,v}$ list will be
empty. Note that the space needed for the support lists is only
$O(|E|)$.

\subsection{Performing \proc{Remove}}
A remove operation is performed by, for each single assignment
$(x_i,v)$ to be removed, visit all nodes that offer support for
$(x_i,v)$. On each such node the update procedure \proc{RemoveEdge} is
used to remove the corresponding edge while maintaining $S$ and the
valid domains. Both are shown in Figure \ref{code:inc_remove}.

\comment{
\begin{lemma}
Executing \proc{Remove} on an MDD $F$ makes $F$
\end{lemma}
\begin{proof}
Proof can be derived directly from \cite{REGULAR-CONSTRAINT}.
\end{proof}}

\begin{figure} \label{code:inc_remove}
\begin{codebox}
\Procname{$\proc{Remove}(R)$}
\li \For each $(x_i,v) \in R$
\li \Do \For each $(u,c) \in s_{i,v} $
\li $\proc{RemoveEdge}(u,c,v)$
\End
\End
\end{codebox}
\begin{codebox}
\Procname{$\proc{RemoveEdge}(u,c,v)$}
\li $C_u \gets C_u \setminus \{ (c,v) \}$
\li $P_c \gets P_c \setminus \{ (u,v) \}$
\li $s_{l(u),v} \gets s_{l(u),v} \setminus \{(u,c)\} $
\li \If $s_{l(u),v} = \emptyset$
\li \Then $D_{l(u)} \gets D_{l(u)} \setminus \{v\}$
\li \If $D_{l(u)} = \emptyset$
\li \Then \Return Constraint failed
\End
\End
\li \If $C_u = \emptyset$ \RComment{$u$ dies a \emph{NoValue} death }
\li \Then \For each $(p,v') \in P_u$
\li \Do $\proc{RemoveEdge}(p,u,v')$
\End
\End
\li \If $P_c = \emptyset$  \RComment{$c$ dies a \emph{NoReference} death}
\li \Then \For each $(c',v') \in C_c$
\li \Do \proc{RemoveEdge}$(c,c',v')$ 
\End
\End
\end{codebox}
\caption{\proc{RemoveEdge} takes as input the dead edge in form of the
  originating node $u$, the destination node $c$ and the corresponding
  value label $v$. It then moves from the dead edge downwards in depth
  first manner as long as there are nodes dying because of lack of
  references. If the node $u$ has no more outgoing edges it propagates
  upwards, removing nodes that have no more outgoing edges. Note that
  any call to \proc{RemoveEdge}, except the initial one by
  \proc{Remove} will only result in \emph{either} upwards or downwards
  propagation. It is not possible to do both. This is because
  downwards propagation is only caused by nodes with no incoming
  edges, while upwards propagation is only caused by nodes with no
  outgoing edges. Furthermore the direction will remain the same after
  the initial invocation by \proc{Remove}.}
\end{figure}

\subsection{Complexity}
We start by examining the complexity of \proc{Remove} in terms of the
number of calls to \proc{RemoveEdge}. Ideally we would like an
algorithm that never uses more time than the scanning approach per
step, and gives an improved bound on the time spent in total. Finally
we show how to choose data structures to support \proc{RemoveEdge} in
time $O(1)$ per call.

\subsubsection{Worst case performance in a single step}
Recall that a \emph{step} corresponds to a request to apply
generalized arc consistency which again corresponds to a call to
\proc{Remove} with a set of assignments banned by other
constraints. The following result bounds the complexity of the
\proc{Remove} procedure in a single step.

\begin{lemma}\label{inc_worstcase}
  The number of calls to \proc{RemoveEdge} in the \proc{Remove} in
  step $t$ is $|E^t_{ref} \cup E^t_{val}|$
\end{lemma}
\begin{proof}
Consider the $\proc{RemoveEdge}(u,c,v)$ method. Assume it is called
with an edge that has just died. Its then easy to see that it will
only invoke \proc{RemoveEdge} on edges that die as a consequence of
the initial call. Since the first invocation of \proc{RemoveEdge} in
\proc{Remove} is guaranteed to be on a dead edge, and since
\proc{RemoveEdge} maintains $S$, we can conclude that
\proc{RemoveEdge} is only called on newly dead edges.
\end{proof}

The dynamic algorithm can potentially use more time in a single step
but we can give a (pessimistic) bound on how much slower the dynamic
approach could be in total.

\begin{lemma}\label{lemma:worstcase-complexity}
If \proc{RemoveScan} visits $s$ edges in any given step $t$, then
\proc{Remove} causes at most $O(s)$ calls to \proc{RemoveEdge} in step
$t+1$.
\end{lemma}
\begin{proof}
This is easily seen by the fact that any edges that die in a step,
must have been alive in the previous step. Combining this with Lemma
\ref{fullscan_complexity} and Lemma \ref{inc_worstcase} yields the
claim.
\end{proof}

Since there can potentially be a factor of $d_{max}$ more search nodes
at step $t+1$ compared to step $t$, the above result means that the
dynamic algorithm might theoretically work on a factor $d_{max}$ more
edges in total. This is a natural consequence of the fact that
\proc{RemoveScan} (in its best case) looks only at living edges, while
the dynamic algorithm spends time on edges that died in the current
step. 

\subsubsection{Complexity over a search path}
Consider a path in the search tree implicitly represented by the
branching search. We will here and in the following describe the
complexity of the presented algorithms as the complexity over such a
root-to-leaf path in the search tree. The following result tells us
that the number of \proc{RemoveEdge} operations required is only
linear in the size of the MDD. As comparison the scanning approach
could use on the order of $t|E|$ operations over a search path with
$t$ steps.

\begin{lemma}\label{lemma:path-complexity}
Consider any root-to-leaf path in the search tree. Then the total
number of calls to \proc{RemoveEdge} is at most $O(|E|) $
\end{lemma}
\begin{proof}
Follow directly from Lemma \ref{inc_worstcase} as edges can only die
once.
\end{proof}

\subsubsection{Complexity of \proc{RemoveEdge}}
\begin{lemma}
A call to \proc{RemoveEdge} that causes a total of $k$
\proc{RemoveEdge} calls can be performed in time $O(k)$ by choosing an
appropriate representation of the MDD. Furthermore the $j$ edges
supporting an assignment $(x_i,v)$ can be enumerated in time $O(j)$.
\end{lemma}
\begin{proof}
The constant time complexity for \proc{RemoveEdge} is easily achieved
using some simple pointer based data structures.

In each node we store its incoming and outgoing edge lists as double
linked lists and each edge is contained in both its start point's
children list and its end point's parent list. Therefore, given an
edge we can remove it from the MDD in $O(1)$ time.

We store the support lists as double-linked lists. Since the support
list in each entry stores the corresponding edge we spend $O(1)$ time
deleting an edge and its support list entry given the support list
entry. Each edge also stores a pointer back to the entry it
corresponds to in the support lists, ensuring that the we can also
delete the edge in $O(1)$ given its support list entry.

The only further operation required by \proc{RemoveEdge} is iteration
over the set of children and lists, which is of course supported by
the lists in time $O(1)$ per element.

Finally enumerating the $j$ supporting edges of a given set of
assignments takes $O(j)$ time as it corresponds to iterating over the
corresponding support list which are maintained such that they only
contain live support edges.
\end{proof}

In practice the outgoing edges of a node will most likely fit in the
memory cache and hence it could be better to simply mark dead outgoing
edges and scan the edge list instead of storing pointers into it. This
approach goes well in hand with an alternative support caching
strategy where we store a single support for each node and when this
support is removed, simply scan the corresponding layer for a
replacement.  Assuming the outgoing edges of each node fit in cache
this will not increase the asymptotic number of cache misses in total
on a root-to-leaf path. If nodes are stored as an outgoing edge set
and a pointer to the incoming edge set layer by layer it will most
likely improve it in practice. However, it does run the risk of
pushing work down the search tree, as we may iterate over dead nodes.
Similarly incoming edge lists can also be stored as arrays and dead
entries simply marked without affecting the root-to-leaf complexity.
However, in order to ensure the complexity over a root-to-leaf path
each edge must still contain an index into its end node's incoming
edge table, as the incoming edges can not be assumed to fit into
cache.

\subsection{No-good recording}
Just as in \cite{ADHOC-GAC} we can use no-good recording for
constraints $f$ when $|scope(f)| < n$. We could also apply good
recording, but that would mean postponing updates to the data
structure which we prefer to do as early as possible in the search
tree.

We expect the no-good recording to be very beneficial, when it
applies, as it can save the potentially costly operation of having to
delete all the remaining edges in the MDD. However, our approach of
attempting to compile as many constraints as possible into a single
MDD constraint could easily result in a scope that contains all
variables.

\section{Skipping input variables} \label{sec:long-edges} 
\begin{figure}\label{fig:edgetypes}
\includegraphics[height=5cm]{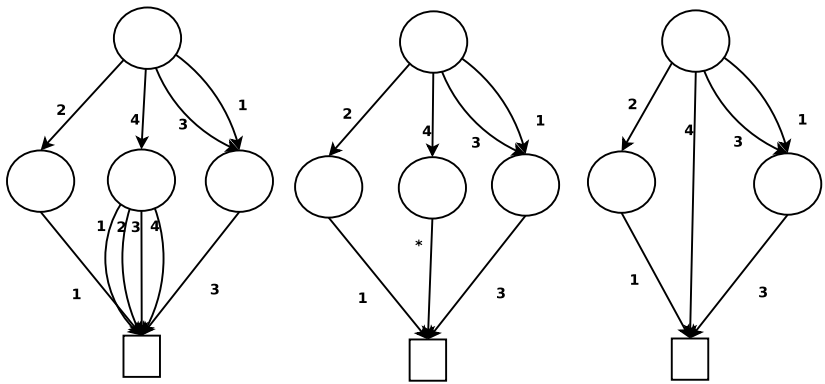}
\caption{The decision diagrams above illustrate the effect of various
  methods for skipping input variables. The leftmost MDD does not skip
  variables, the middle one uses wildcard nodes, and the rightmost
  uses long edges.}
\end{figure}

We have so far obtained a very efficient GAC algorithm for a
simplified MDD data structure. In particular the algorithm described
so far does not allow the MDD to be fully reduced. While simple and
efficient to use, the simplified MDD is not as compact as a fully
reduced MDD. It is inefficient in the case where a node $u$ exists in
the MDD representing a choice point that has no effect, ie all choices
lead to same node $c$. Recall that in a fully reduced MDD such a node
would have been removed and its parent would instead point directly to
$c$ (if no parent exist, $c$ becomes the new parent and we have an
implicit edge skipping layer 1 to $l(c)-1$). Such an edge that skips a
layer in the MDD is called a \emph{long edge}. In the extreme the
difference in edge count between the simplified MDD and the fully
reduced MDD can be a factor of $nd_{max}$. An alternative that is
sometimes used is to introduce wildcard nodes instead of long edges. A
wildcard node only have one outgoing edge, indicating that all the
node's $d_i$ edges point to the same end point. This can only yield a
factor of $d_{max}$ reduction in edge count over the simplified MDD,
but the changes required to support wild card nodes are simpler than
for long edges. Figure \ref{fig:edgetypes} illustrates the different
edge types.

\subsection{Handling long edges} \label{subsec:long_edges} 
The simplest way to support long edges is to simply expand them for
the purpose of building support lists. However, this means we use time
in the total length of all the long edges, so we can no longer provide
a complexity bound linear in the number of edges over a root-to-leaf
path in the search tree.  In the scan approach the issue of long edges
is solved by scanning the BDD and for each level finding the longest
outgoing edge. Given this information it is simple to list the
variables which have full support due to a long edge in time $O(n)$
\cite{VD-NOTE}.

In our dynamic approach we do not wish to scan the MDD in order to
find the longest edges. Instead, for each distinct interval supported
by at least one long edge, we store the counter $L_{i,j}$, being the
size of the set of long edges skipping the layers $i$ to $j$.

We will maintain the set $\mathcal{L}$ of these intervals during the
search and based on these decide which variables are supported by long
edges. A long edge $e$ dies if its end node $d(e)$ dies or if the
assignment $(x_{s(e)},v(e))$ becomes invalid. Note that we ignore the
case where a long edges dies by having all values for one of its
skipped variables removed. This is safe because in this case the
constraint fails. This also means we do not keep track of how many
values are actually available for each skipped variable or modify the
intervals if a \proc{Remove} call actually 'cuts' a long edge into two
parts. This means that we allow the intervals to support values that
are invalid. However, the only invalid values covered in this way are
those that have explicitly been removed by calls to \proc{Remove} so
we can easily correct this by modifying \proc{Remove} to eliminate the
domain values corresponding to its arguments as we only perform
decremental updates to the domains after that.

Each long edge $e$ skipping the layers from $i$ to $j$ stores a
pointer to $L_{i,j}$ and when $e$ dies we decrement $L_{i,j}$. When
the counter for an interval reaches zero, the interval is no longer
supported by any long edge and is therefore removed from
$\mathcal{L}$. We can easily create a dynamic version of the technique
applied to the scanning approach for BDDs in order to handle long
edges in \cite{VD-NOTE}. Recall that in this case a table listing the
longest outgoing edges originating from each layer is computed, and
based on this the variables covered by long edges can be computed in
time $O(n)$. We can obtain a dynamic version of this solution approach
by store a priority queue for each level, storing the longest interval
starting at that layer. Assuming the priority queue supports reporting
the maximum in time $O(1)$ we can obtain the required table in $O(n)$
time as long as the priority queues are properly maintained which is
done in the same manner as in the previously described approach. The
following lemma gives the amortized complexity of this algorithm.


\begin{lemma}
On a root-to-leaf in the search tree the complexity of the
longest-outgoing-edge based approach is bounded by $O(|E| + n^2 \lg\lg
n + n^2d_{max})$.
\end{lemma}
\begin{proof}
  As previously, the actual time for handling normal edges is at most
  $O(|E|)$. Each interval can only be removed once, and each such
  deletion costs $O(\lg\lg |\mathcal{L}|)$ using a VEB-based priority
  queue\cite{VEB}. Finally we spent $O(n)$ time per step to compute
  the table of longest outgoing edges and compute the variables
  covered by long edges. In total this yields a complexity over a
  root-to-leaf path in the search tree of $O(|E| + |\mathcal{L}|\lg\lg
  |\mathcal{L}| + tn)$, where $t$ is the number of steps. Since there
  can at most be $O(n^2)$ distinct long edge intervals and $nd_{max}$
  steps this yields $O(|E| + n^2 \lg\lg n + n^2d_{max})$.
\end{proof}

As an alternative solution we can use the dynamic interval union data
structure (DIU) presented in \cite{DYN-INTERVAL-UNION} to store the
intervals. The DIU allows us to add or remove intervals in time $O(\lg
|\mathcal{L}|)$ while allowing enumeration of the $k$ disjoint
intervals representing the union of the stored intervals in time
$O(k)$. Furthermore the list of values lost from the domain can be
computed during a delete using just the time to enumerate them. This
approach yields the following result.

\begin{lemma}
On a root-to-leaf in the search tree the complexity of the DIU based
approach is bounded by $O(|E| + n^2\lg n + nd_{max})$.
\end{lemma}
\begin{proof}
As previously the time spent on deleting edges and handling normal
edges is at most $O(|E|)$. Each interval can only be removed once and
each such removal takes time at most $O(\lg |\mathcal{L}|)$. The only
other work performed is to record the value removed in each step,
failing the constraint if needed, costing $O(1)$ per step. This leads
to a complexity of $O(|E| + |\mathcal{L}|\lg |\mathcal{L}| +
t)$. Since there can at most be $O(n^2)$ distinct long edge intervals
and $nd_{max}$ steps we obtain $O(|E| + n^2\lg n + nd_{max})$.
\end{proof}

We believe that for most practical applications of the MDD constraint
the above complexity will be completely dominated by the $|E|$ factor,
and hence that the addition of long edges result in no significant
performance impact. The DIU approach is most likely preferable to the
longest-outgoing-edges approach in practice unless $n$ is very large
and the domains very small, but replacing the VEB with a standard
binary heap yields a very simple approach with a decent asymptotic
complexity and good practical performance.

\subsection{Handling wild card nodes}
Instead of removing the nodes in a long edge, we can replace each of
them with a wild card node. A wild card node has a single edge
labelled $*_i$ representing all the outgoing edges. An interesting
point about such an edge, is that during calls to \proc{Remove} or
\proc{RemoveScan} the meaning of what allowed values it corresponds to
may change, but it will change to the same for all edges labelled with
$*_i$. Therefore, we never need to update the outgoing edge of a
wildcard node, just as we didn't update long edges. When the scanning
approach is used and we encounter a wild card node $u$ which points to
a live node, we can add all values in $D_{l(u)}$ to the valid domains.

When using the dynamic approach the wild card edges are not represented
in the support lists. Instead a separate table containing a count of
wild card nodes $w_i$ for each level is stored along with a list of
domain values $s^w_{i}$ that can only be supported by wild card nodes.
When a wild card node $u$ dies, $w_i$ is decremented, should this yield
$w_i = 0$, the values in $s^w_i$ are removed from $D_i$. Finally, when
an assignment $(x_i,v)$ loses support in the support list,
\proc{RemoveEdge} now only removes $v$ from $D_i$ if $w_i = 0$ and
adds $v$ to $s^w_i$ otherwise.

 \section{Maintaining the reduced property} \label{sec:reduce} 

 In the above as well as in \cite{ADHOC-GAC} we do not take steps to
 maintain the uniqueness reduced property of the MDD when we update
 the data structure. This forfeits a chance for a large speed-up. If a
 reduction at an early search node $s$ would lead to large reduction
 in the size of the data structure all descendant search node of $s$
 (of which there can be an exponential number) would benefit from
 working on a much smaller data structure. An example showing the
 effect of dynamic reduction is given in Example
 \ref{ex:reduction_benefit}.

\begin{example}\label{ex:reduction_benefit}
  As an example of the effect of dynamic reduction consider the simple
  constraint encoding the rule $x_1 \le x_2, x_1 \le x_3 \ldots, x_1
  \le x_j$ with domains $D_i = \{1, \ldots, k \}$ for some constant
  $k$. Let $f_v$ denote the sub-structure representing the constraint
  restricted to $x_1 = v$.  Now consider the removal of the value $1$
  from the domain of variable $x_2,\ldots, x_j$ (as could be induced
  by an external AllDifferent constraint). With this restriction $f_1$
  becomes equivalent to $f_2$ and can be merged, reducing the size of
  the MDD with a constant factor. If the value $2$ is lost next then a
  further constant fraction of the MDD can be removed due to the
  reduce step as $f_1 = f_2$ now becomes equivalent to $f_3$. This is
  of course a very simplistic constraint easily propagated using other
  methods, but if we consider the conjunction of the constraint with
  another constraint, the example still applies in many cases,
  especially if the new constraint does not depend on the value of
  $x_1$. One example of such an additional constraint is $\forall i
  \in [2,j-1] : x_i \not = x_{i+1}$.
\end{example}

As we will see below, the scanning approach can be easily adapted to
perform reductions, though $\Delta$-cutoff loses its benefits. The
dynamic approach incurs a small performance penalty, but still
elegantly avoids falling back on a scanning approach. Note that if we
ensure the uniqueness reduced property the MDD will be fully reduced
according to the original domains throughout the search (assuming it
is fully reduced initially), since there is no risk of new long edges
when we only perform domain restrictions. We therefore first discuss
how to ensure the uniqueness reduced property and in section
\ref{subsec:full_reduce} describe the addition necessary in order to
obtain full reduction according to the \emph{current} domains. In
section \ref{subsec:domain_entailment} we cover domain entailment
detection for fully reduced and uniqueness reduced MDDs.

\subsection{Static reduction}\label{subsec:static_reduction}
Our goal is to provide reduction along with our dynamic generalized
arc consistency algorithm. However we will first cover how to reduce
the MDD statically, such as could be done in conjunction with
\proc{RemoveScan}. To that end we make the following observation:

A node $u_1$ in layer $i$ can become redundant iff the death or merger
of one or more of its children renders its live outgoing edge-set
$C_{u_1}$ identical to another node (assuming that the MDD is
uniqueness reduced for all layers below layer $i$). In order to remove
this redundancy $u_1$ and $u_2$ should merge, by letting $u_1$ be
subsumed by the identical node $u_2$(in this case we say that $u_1$ is
the subsumee and $u_2$ the subsumer), or vice-versa.  Assume that the
MDD is reduced for all layers below layer $i$. Given a node $u_1$ in
layer $i$ which changed its outgoing edges, we merely need to check if
there exists another node on level $i$ which have the exact same set
of children.  If such a node $u_2$ is found, $u_2$ subsumes $u_1$ by
redirecting all incoming edges that end in $u_1$ to $u_2$ and deleting
all outgoing edges of $u_1$.  Following this reduction each modified
parent must be tested for redundancy. To ensure the reduced property
for lower layers at all times the scan can be changed to operate in a
breadth first manner, postponing all reductions until reaching the
lowest layer affected by \proc{RemoveScan} at which point the
reductions can proceed in a bottom-up manner.

For the uniqueness test we use hashing, each node $u$ is hashed as the
pair $(C_u,l(u))$ and inserted into a hash table, updates being
performed by recomputing the hash value and reinserting the node. The
cost of performing the reduction during step $t$ is therefore expected
$O(|E_{liv}^{t-1}|)$ assuming that universal hashing is used. The
space used for the hash table is $O(|V|)$ and will therefore not have
a significant impact on the total space used compared to the space
needed to store the edges of the MDD.

Note that $\Delta$-cutoff is no longer useful as we have to scan the
entire live part of the structure in order to ensure that its reduced.

\subsection{Dynamic reduction}
Let us start by considering an adaption of the static approach
described above. To obtain a dynamic version we will resolve detected
redundancies as above by subsuming one node into another. We can also
use the same redundancy hash table. Instead of scanning for redundant
nodes we simply need to check nodes that lose an outgoing edge, or
have an outgoing edge redirected due to a subsume operation.

In order to make this approach correct and efficient there some issues
which must be addressed. The first is to ensure that subsumptions and
redundancy checks happens in the correct order. The second issue is
that the process of subsuming a node can potentially be very expensive
since all edges pointing to the subsumee needs to be moved. Finally we
need to be able to quickly perform a redundancy check on a node, it is
no longer acceptable to spent time linear in the number of outgoing
edges for each such check.

\subsubsection{Ordering}
To ensure that nodes are considered for reduction in the correct
order, we maintain a set of 'dirty' nodes that need to be checked for
redundancy. When the removal phase ends we can check these for
redundancy in a bottom-up manner. Note that a redundancy check in
layer $i$ can lead to a subsumption, which can lead to redundancy
checks and subsumptions in layer $j < i$, but not in layer $j' > i$.
If we do not wish to reduce at every step it is safe to maintain the
set of 'dirty' nodes between steps.

\subsubsection{Redundancy detection}
As in the static reduction approach we will use a hash table to check
for redundancy. However we need a hash value for each node that can be
updated in $O(1)$ time when an outgoing edge is lost or updated.
Furthermore, when inserting nodes into a hash table, it might be that
other nodes have hashed to the same location, either due to a hash
collision or due to the inserted node being redundant. If we do not
have any efficient way to checking if it is a collision or not we will
need to compare the inserted node $u_1$ with each of the collided
nodes $u_2$ in turn. Such a comparison requires us to check whether
$C_{u_1} = C_{u_2}$, taking time $O(\min \{ |C_{u_1}|,C_{u_2}| \})$,
and hence an insertion could require time $O(d_{max})$ for a constant
number of collisions. We therefore require an approach that can ensure
that we (almost) never need to do a full comparison with another node
unless that node makes the inserted node redundant.  Such an approach
will ensure that insertions only take $O(1)$ time per collision. In
case the inserted node is redundant we will need to perform one full
comparison. To resolve such a redundancy we need to remove all edges
of one of the nodes anyway, so the asymptotic complexity is not
affected. For now we will assume the availability of a hashing
strategy with the above properties and show what can be achieved under
this assumption.  Afterwards we show how to achieve such a hashing
strategy in practice.

\subsubsection{Merging nodes}
Given two identical nodes $u_1$ and $u_2$ to merge we always designate
the one with the largest number of parents as the subsumer in order to
reduce the total cost of the merge operations. An edge $e$ is only
moved when its end-point $c$ is subsumed. Since this only happens when
another node of larger in-degree becomes identical to $c$ the
in-degree required to cause $e$ to be moved must at least double each
time $e$ is moved. Hence an edge can only be moved $\lceil \lg(|V|)
\rceil$ times as $|V|$ is an upper bound on the in-degree of a node.
Note that this is a very simple and classic greedy strategy that
incurs no significant overhead.

\subsubsection{Complexity of dynamic reduction}
\begin{lemma}
  Let $m_i$ be the number of layer $i$ edges that are incident to a
  node involved in a subsumption during the part of the search
  corresponding to a given root-to-leaf path in the search tree. The
  time spent over this path by \proc{Remove} on reducing the MDD is
  then $$O(\lg |V| \sum_{1\le i \le n}m_i) = O(|E|\lg |V|)$$
\end{lemma}
\begin{proof}
  Outgoing edges of nodes that are subsumed are simply deleted,
  requiring $O(1)$ time per outgoing edge. Note that each end point of
  the deleted edges much have an in degree of at least two before the
  subsumption, and therefore no further edges will need to be removed.
  In total this sums to $O(m_i)$. Moving an incoming edge from the
  subsumee to the subsumer takes $O(1)$ time using the above mentioned
  hashing strategy. As demonstrated earlier, each edge will be moved
  at most $\lceil \lg |V| \rceil$ times. Therefore, we find that the
  total time spent on reducing the MDD is $O(\lg(|V|)\sum_{1\le i \le
    n}m_i)$.
  
\end{proof}

Note that all of the above techniques could be applied to the scanning
approach but the asymptotic performance would not change in the worst
case. We also note that this dynamic approach will never use
asymptotically more time on merging in total or in a single step
compared to the previously described scanning approach to reduction.

\subsubsection{Hashing strategy}
In order to fulfill the required promises for the hashing strategy we
use the following two techniques.

\paragraph{Fast updates}
To allow quick updates of hash values we will use a slight variation
on vector hashing\cite{universal-hashing,VECTOR-HASHING}. The idea
behind vector hashing hashing is to hash a vector by assigning one
hash function to each entry in the vector. The hash value is then
computed as the XOR of each entry's hash value. The most interesting
property of vector hashing is that if a single entry changes value,
the hash value can be updated in constant time, regardless of the
length of the vector. In order to apply vector hashing it is necessary
that each entry's hash function is chosen from a \emph{strongly}
universal class of hash functions:

\begin{definition}[Strongly universal hashing]\cite{VECTOR-HASHING}
  A class of hash functions $\mathcal{H} \subseteq (U \rightarrow U)$
  is said to be strongly universal if for all distinct $x,y \in U$ it
  holds that $\forall \alpha,\beta, \alpha \not = \beta: \Pr [h(x) =
  \alpha \land h(y) = \beta] = O(1/|U|^2)$ for any $h$ chosen
  uniformly randomly from $\mathcal{H}$.

\end{definition}

Our intention is to hash a node $u$ as a bit vector $b$ of length
$d_{max}+1$ such that for $j \le d_{max}$ the $j$th entry is the index
of the node that is the end-point of the outgoing edge labelled $j$ if
it exists, and $0$ otherwise. For the remaining entry we put $b_{d+1}
= l(u)$, to distinguish between identical nodes in different layers.
Note that this is just an encoding of $(C_u,l(u))$ which was also used
as key in Section \ref{subsec:static_reduction}. If we apply vector
hashing directly we will need time $O(d_{l(u)})$ per node to compute
the initial hash values, the total cost of which might not be bound by
$O(|E|\lg|V|)$.  With a slight modification presented in Lemma
\ref{lemma:set-hashing} the total time required for computing the
initial hash values is reduced to $O(|E|)$.

\begin{lemma}\label{lemma:set-hashing}
  Let $\mathcal{H} \subseteq (U \rightarrow U)$ be a class of strongly
  universal hash functions. Let $U_0$ denote a chosen 'null' element
  of $U$. Define $\mathcal{H}_{U_0}^S$ as the class of hash functions
  $h : U^d \rightarrow U$ of the form $h(u) = h_0(U_0) \oplus (
  \oplus_{u_j \not = U_0} h_j(u_j) )$ where $\{ h_0, \ldots, h_d \}
  \subseteq \mathcal{ H}$. Then $\mathcal{H}_{U_0}^S$ is strongly
  universal.
\end{lemma}
\begin{proof}
  Consider two vectors $u = (u_1, \ldots, u_d)$ and $q = (q_1, \ldots,
  q_d)$ such that $q \not = u$.  We need to show that $\forall
  \alpha',\beta',\alpha' \not = \beta' : \Pr[h(q) = \alpha' \land h(u)
  = \beta'] = O(1/|U|^2)$. We consider two cases:

  \paragraph{Case 1}
  First if there is at least one entry $j$ such that $u_j \not = q_j$
  and $u_j,q_j \not = U_0$, consider fixing all hash functions except
  $h_j$.  We then have $h(u) = \alpha \oplus h_j(u_j)$ and $h(q) =
  \beta \oplus h_j(q_j)$ for some fixed $\alpha$ and $\beta$.

  We note that for a given choice of $\alpha, \beta, \alpha',\beta'$
  there exists at most one pair of values for $h(u_j)$ and $h(q_j)$
  such that $h(u) = \alpha' \land h(q) = \beta'$. Since $h$ is chosen
  from a strongly universal class of hash functions we therefore have
  $\Pr[h(u) =\alpha' \land h(q) =\beta'] = O(1/|U|^2)$.

  \paragraph{Case 2} 
  If the first case does not apply then for all $1 \le k \le d$ it is
  the case that either $u_k$ or $q_k$ is $U_0$ or $u_k = q_k$ . For
  this second case assume without loss of generality that $u_j = U_0
  \not = q_j$ for some $j$.  We then have $h(u) = \alpha$ and $h(q) =
  \beta \oplus h_j(q_j)$. We note that in choosing a random hash
  function from $\mathcal{H}_{U_0}^S$ we also choose its component
  hash functions independently. Therefore $h_j(q_j)$ is independent
  from $h(u)$.  Furthermore we note that for given values of $\beta$
  and $\beta'$ only one value of $h_j(q_j)$ results in $h(q) =
  \beta'$.  Hence we obtain $\forall \alpha',\beta',\alpha' \not =
  \beta' : \Pr[h(u) =\alpha' \land h(q) =\beta'] = O(1/|U|^2)$.
\end{proof}

To use the above result we choose a $U$ such that $V \subseteq U$ and
set $U_0 = 0$. Note that the addition of $h_0(U_0)$ in the lemma is
only required to ensure strongly universal hashing when $U_0^d$ is
allowed as key.  Since this is not the case for our redundancy checks
($0^d$ corresponds to a dead node) $h_0$ is not needed.

Note that the the initial computation of the hash values during
construction of the MDD now only requires $O(|E|)$ XOR operations, as
the hash values only depend on existing edges. Finally, a hash value
$q$ for a node $u$ can be updated in time $O(1)$ when an outgoing edge
$(c,v)$ is updated to $(c',v)$, simply by computing $q \oplus h_v(c)
\oplus h_v(c')$.

\paragraph{Avoiding unnecessary comparisons}
So far we have not discussed an appropiate size for $U$. In practice
$U$ will be chosen such that $|U| = 2^w$ where $w$ is the word size of
the relevant machine in bits. Because we use strongly universal
hashing, a class of hash functions obtained by truncating the hash
value to a specific length is also strongly
universal\cite{VECTOR-HASHING}. Hence we can generate word size hash
values and use a prefix to index the hash table. We will use the
remaining bits to solve the issue of expensive collisions in the hash
table, in the following way: When it becomes necessary to compare
nodes within a bucket we compare the remaining bits of the hash value,
and only if these are identical will we perform a full comparison of
the two nodes in question.

To analyze the performance of this approach let us assume that the
generated hash values contain $\lceil \lg k \rceil$ bits more than
required to index the hash table, for some $k$. At any point during
the search at most $|V|$ nodes are present in the hash table. In total
at most $O(|E|\lg|V|)$ different nodes will be inserted during the
processing of a root-to-leaf path in the search tree, since each node
updates it hash value each time one of its outgoing edges is updated
or removed.  The expected number of elements per bucket in a hash
table is $l_b = O(1)$.  The expected number of nodes with the same
full hash value is $l_h = O(1/k)$. Each insertion costs $O(l_b +
d_{max}l_h)$.  Hence the expected cost of each insertion, of which
there is at most $O(|E|\lg|V|)$, is $O(1 + d_{max}/k)$.  If $k =
\Omega(d_{max})$ we obtain a total expected cost of $O(|E|\lg|V|)$ for
insertions.

\subsection{Full reduction based on current domains}\label{subsec:full_reduce}
The reduce step described above keeps the MDD fully reduced according
to the original domains. This means that while the MDD is uniqueness
reduced it is not fully reduced according to the \emph{current}
domains. As an example consider an MDD with 1 variable $x_1$ and a
single node $u_1$ with edges $1$ and $2$ going to \textit{terminal}.
If the domain of $x_1$ is $\{ 1,2,3 \}$ this MDD is fully reduced,
while it reduces to the terminal node if the domain is $\{1,2\}$.

Full reduction according to the current domains can be achieved by
using the following rule: If a node $u$ has live edges with labels
corresponding to all values in $D_{l(u)}$ to the same child we will
consider it redundant and reduce it into a long edge (or wild card
node). Note that this will not result in incorrect values being added
to the domains as observed in Section \ref{sec:long-edges} and that
this edge can only lose its supporting values if the corresponding
domain is empty, in which case all the constraints fail so again we do
not need to keep track of the actual 'content' of the edge.

However, in order to maintain the MDD reduced under these new rules,
we will after a domain value for $x_i$ is removed need to discover
nodes that now support all possible values of $x_i$ while only having
a single distinct child node.

First off we need to be able to efficiently discover that a node only
has one distinct child. To this end we observe that a node either
starts out with all edges pointing to the same node or achieves this
status through loosing outgoing edges or having two child nodes merge
to one. We can handle the second case by for each node either
maintaining a unique child hash table or by using a counter in
conjunction with maintaining links between edges leading to the same
node.

When a node is discovered to only have one distinct child node we
store them in a hash table in a way that allows us to retrieve them
based on the values they support. We simply maintain a hash signature
for each node on the set of labels in use on its outgoing edges, using
the same variation of vector hashing used earlier, this time treating
a set of value labels as a bit vector with $d_{max}$ entries. When a
node is discovered to only have one distinct child node, we insert it
into a hash table using the above mentioned signature. We also
maintain this hash signature for each domain. When a domain value is
lost, we merely update the domain signature and look up all edges
corresponding exactly to the current domain (nodes having a super set
of the current domain will have been discovered in an earlier step).

None of this affects the asymptotic space usage or amortized
complexity of the previous reduction technique. We note that in
practice we can combine the hash table needed for the uniqueness
reduction and the one needed for full reduction under current domains
into one in order to save space.


\subsection{Domain entailment detection}
\label{subsec:domain_entailment} Given a constraint $F_k$, and a
partial assignment $\rho$, let $sol_\rho(F_k)$ be the set of vectors
of domain values corresponding to solutions allowed by this constraint
that are consistent with $\rho$. A constraint $F_k$ is said to be
\emph{domain entailed} under domains $D$ iff $\times_{1 \le i \le n}
D_i \subseteq sol_{\rho}(F_k)$\cite{vanhentenryck94design}. That is,
if all possible solutions to the CSP based on the available domains
will be accepted by $F_k$ then $F_k$ is entailed by constraints
implicit in the domains. It is beneficial to be able to detect domain
entailment as it allows the solver to disregard the entailed
constraint until it backtracks through the search node where the
constraint was entailed.

If an MDD is kept fully reduced according to the current domains it is
entirely trivial to detect domain entailment as the MDD will be
reduced to the terminal node.

If the MDD is only kept uniqueness reduced and is domain entailed it
is easy to see that it will consist of a path of precisely $n+1$
nodes(incl. the 1 terminal) if we use wild card nodes and a path of up
to $n+1$ nodes if we use long edges. Note that this state of the MDD
is both necessary and sufficient for domain entailment assuming that
the MDD has performed the most recent domain propagation step. We can
easily track whether or not the above properties are fulfilled using
the following rules: If there is at most 1 live node per layer and the
MDD constraint has not failed it is domain entailed. Naturally the
node count can be maintained efficiently by simply updating a layer
node counter whenever a node dies and maintaining a further counter
for the number of layers having a node count of 1 or less.

\section{Constructing the MDD}\label{sec:construction}
In order to apply our approach we first need to construct the MDD and
compute the necessary auxiliary data structures.

The input to constructing the MDD is assumed to be a set of
constraints expressed in discrete variable logic. For example, tabular
constraints could be expressed as disjunction of tuples, while an
AllDifferent would be expressed as $\forall (x_i,x_j) \in X^2, i \not
= j : x_i \not = x_j$.

We suggest to construct the MDD by first building the ROBDD of the
component constraints using $\lg d_i$ binary variables to represent
domain values for $x_i$(see for example \cite{BDD-CONFIG}). This
allows utilization of the optimized ROBDD libraries available and
furthermore gives access to the many variable ordering heuristics
available for BDDs which can substantially reduce the size of the BDD.

After the ROBDD is constructed it is trivial to construct the MDD from
the BDD using time linear in the resulting MDD, assuming the binary
variables encoding each domain variable are kept consecutive in the
variable ordering.  The additional data structures required by the
incremental algorithm can be obtained by using the scanning approach
to discover all the supporting edges.


The time that is acceptable for the compilation phase(and therefore
also the allowable size of intermediate and final MDDs) depend on
whether the constraint system is to be solved once or whether it is
used in for example a configurator where the solver is used repeatedly
on the same constraint set(with different user assignments) to compute
the valid domains\cite{CONFIG-COMPARE}. One could easily specify a
large set of constraints and incrementally combine them into fewer and
fewer MDD constraints until a time or memory limit is reached and
still gain the benefit of improved propagation.

\section{Other constraint compilation data structures}\label{sec:other_compilation}
While we have described the above algorithms in terms of MDDs our
approach also applies to similar data structures as described below.

\subsection{Interval edges}\label{subsec:interval-edges}
In an ordinary MDD each edge corresponds to a single domain value. It
is quite natural to consider the generalization to edges that
represent a subset of a domain. One particular useful generalization
of edges is to let each edge correspond to some interval of the domain
values. This approach is used in Case
DAGs\cite{SICTUS-PROLOG-MANUAL,BCC} which resemble MDDs without long
edges, but where edges represent disjoint intervals instead of single
values.

We could directly apply our approach above to such a Case DAG, but
nodes might be stored in many more support lists than they have actual
edges. This can be fixed quite easily however.

\subsubsection{Basic idea}
For each unique interval $I^j \in \mathcal{I}_i$ in layer $i$ we store
a set $I^j_E$ of edges labelled with this interval. To each such
interval we associate a counter $I^j_c$ specifying the number of valid
values in the interval(so initially $I^j_c$ will simply be the size of
the interval). We use $l^I$ to denote the sum of the size of all
distinct interval present in layer $i$ and note that $l^I = O(n
d_{max}^3)$ since there can at most be $O(d_{i}^2)$ distinct intervals
of length at most $d_i$ in layer $i$.

The idea is now to maintain the set of live intervals for each layer.
If we can do that efficiently we just need to be able to detect loss
of domain values, which corresponds to maintaining the union of a set
of intervals efficiently under deletion. Note that we do not need to
split intervals when a \proc{Remove} call splits the allowed domain on
an edge, for similar reasons as described in Section
\ref{subsec:long_edges}.

This idea enables us to create a GAC that algorithm that spends time
depending on the length of the distinct intervals and not on the total
sum of all intervals. As comparison, the straightforward scanning
approach will use time in the total interval size summed over
\emph{all} edges alive in the MDD in each step, meaning that
processing a single edge could cost as much as $d_{max}$.

\subsubsection{Maintaining the live intervals}
In order to maintain the set of live intervals we need to maintain
$I_c$ and $I_E$ for each interval. When performing \proc{Remove} in
layer $i$ we can by using an interval tree find the intervals that
cover the value $v$ and decrement the corresponding counters in time
$O(\lg |\mathcal{I}_i| + k)$ where $k$ is the number of intervals
intersecting $v$. The necessity of the counter on each interval means
that each interval can be accessed by by \proc{Remove} as many times
as the size of the interval of values it supports. \proc{RemoveEdge}
can work as before, the only addition being to update $I_E$ of the
interval associated to the edge being removed.Using this approach the
time for maintaining the live intervals over a search path with $t$
steps is $O(t\lg(|\mathcal{I}|) + l^I + |E|) = O(nd_{max}^3 + |E|)$.

\subsubsection{Maintaining the union}
Since we are already spending time in the total length of the distinct
intervals, we can use a very simple approach to maintain the union of
the intervals. For each layer $i$ we maintain a counter indicating the
number of intervals covering it. When an interval is deleted it
decrements all the corresponding counters. Combined with the
maintenance of the live intervals we still get a complexity of
$O(nd_{max}^3 + |E|)$. The disadvantage of this simple approach is
that these relatively expensive deletions occur at lower levels in the
search tree than the counter decrements. As an alternative we can use
the DIU data structure described in Section \ref{subsec:long_edges} to
maintain the union. Over a search path each interval can be deleted
once so we get a complexity of $O(|\mathcal{I}|\lg(|\mathcal{I}|)) =
O(nd^2_{max}(\lg(n) + \lg(d_{max} )))$ for handling the deletions in
the DIU data structure. The reporting of values no longer covered is
in total $O(nd_{max})$. Adding this to the time complexity for
maintaining the live intervals we obtain a total time complexity of
$O(nd^3_{max} + nd_{max}^2\lg (n) + |E|)$.

\subsection{AND/OR Multi-Valued Decision Diagrams (AOMDD)}
AOMDDs were introduced in \cite{AOMDD}, and from the perspective of a
GAC algorithm just introduces AND nodes into the MDD, such that each
child of an AND node roots an AOMDD which scope is disjoint from its
siblings. This data structure is potentially much more compact than an
MDD. Fortunately the above described technique can be applied
easily. The only change is that an AND node dies if it looses any of
its outgoing edges as opposed to all outgoing edges for an OR
node. Therefore we can utilize this more compact type of decision
diagram while still maintaining the complexity bounds in terms of the
\emph{size} of data structure.

\comment{
\subsection{Tabular constraints}
The concept of caching support of assignments also applies well to a
constraint that is represented simply as a table of acceptable
tuples. The support lists can be created as before, each entry
pointing to a supporting tuple. Assuming that the space used by each
value in a tuple requires at least as much space as a pointer, the
support lists do not increase the asymptotic space consumption of the
tabular constraint. If $t$ is the number of tuples, this achieves an
$O(t)$ bound on the number of tuples examined over any one path in the
search tree. A similar alternative strategy relying on storing just
one support per assignment and searching for a new support only when
the old support is lost is described in \cite{WATCHLIT-MINION}.
}

\section{Future work}
One possible weakness of our approach lies in the inability to share
work between multiple constraints. Just as in \cite{ADHOC-GAC}
identically structured constraints on different scopes can share the
no-good cache, but that is the limit of co-operation. Specifically MDD
substructure sharing between separate constraints does not extend to
the support lists, while the scanning algorithm can share pruned
nodes/edges and the $\Delta$-cutoff value.

Another interesting subject is the propagation between MDD
constraints. Due to the size of the conjunction of a set of
constraints it might be more practical to use a small set of MDD
constraints each being the compilation of a subset of the original
constraints. In this case it might be beneficial to consider a
stronger propagation than just domain propagation among the MDD
constraints. This stronger propagation could for example take the form
of exchanging binary decompositions between constraints, as
projections of the solution space is easy to compute in an MDD.

\section{Conclusion}
This paper introduced the MDD global constraint and provided an
efficient incremental Generalized Arc Consistency algorithm for it
based on techniques from \cite{REGULAR-CONSTRAINT}, that is
potentially much more efficient than the straightforward scanning
approach while not using asymptotically more space. Since the
constraint uses a reduced decision diagram to represent the solution
space of the constraint it can be used to represent tabular
constraints in a compressed manner while still allowing a complexity
that relates to the size of the data structure and not the number of
solutions stored as opposed to normal compression. Furthermore the MDD
global constraint can be used to efficiently represent the solution
space of a set of simpler constraints. As an additional advantage the
constraint can be kept reduced dynamically in an efficient manner while
also allowing efficient domain entailment detection. By using a good
model and a good choice of global constraints, CSPs can therefore be
reduced to a set of normal global constraints and a set of MDD
constraints for improved domain propagation.

\bibliography{bib}
\bibliographystyle{abbrv}
\end{document}